\documentclass[9pt, conference, a4paper, compsocconf]{IEEEtran}
\usepackage{amsmath,amsxtra,amssymb,amsthm,latexsym,amscd,amsfonts}
\usepackage[utf8]{vietnam}
\usepackage[top=2.7cm, bottom = 5.4cm, left=2.72cm, right = 2.85cm]{geometry}
\usepackage{balance}

\usepackage{fancyhdr}
\usepackage{cite}
\usepackage{algorithmic}
\usepackage{textcomp}
\usepackage{booktabs}   
\usepackage{tabularx}  
\usepackage{xcolor}
\usepackage{amsmath,amsxtra,amssymb,amsthm,latexsym,amscd,amsfonts}
\usepackage{pgfplots}
\usepackage{multirow}
\usepackage{subcaption}
\usepackage{url} 
\pgfplotsset{compat=1.17} 

\usepackage{tikz}
\usetikzlibrary{positioning}  
\usetikzlibrary{arrows.meta}  
\usetikzlibrary{shapes}      
\usetikzlibrary{calc}      
\usetikzlibrary{fit}          
\usepackage[english]{babel} 
\usepackage{fancyhdr}
\usepackage{array}

\usepackage{caption}
\usepackage[font=footnotesize]{subfig}

\makeatletter
\def\ps@IEEEtitlepagestyle{%
\def\@oddhead{\hfil \small{\textit{Hội thảo khoa học Quốc gia về Trí tuệ nhân tạo 2026 (FJCAI) - Cần Thơ, 27-28/3/2026}\hfil}%
	\def\@evenhead{\hfil\small{\textit{Hội thảo khoa học quốc gia về Trí tuệ nhân tạo 2026 (FJCAI) - Cần Thơ, 27-28/3/2026}\hfil}}%
		\def\@oddfoot{\scriptsize \thepage \hfil }%
		\def\@evenfoot{\scriptsize \hfil \thepage}
	}
}

\def\@maketitle{%
  \newpage
  \null
  \begin{center}%
    {%
      \fontsize{16}{18}\selectfont
      \bfseries \@title \par
    }%
    \vskip 4em%
    {%
    \fontsize{9}{3.5}\selectfont
      
      \begin{tabular}[t]{c}%
        \@author
      \end{tabular}\par
    }%
  \end{center}%
  \par
}

\renewenvironment{abstract}
  {\normalfont
   \if@twocolumn
     \@IEEEabskeysecsize\bfseries\textit{\abstractname}: 
   \else
     \begin{center}\@IEEEabskeysecsize\textbf{\abstractname}\end{center}
     \@IEEEabskeysecsize
     \setlength{\parindent}{0pt}   
     \noindent
   \fi}
  {\par}

\renewenvironment{abstract}
  {\normalfont
   \@IEEEabskeysecsize
   \setlength{\parindent}{0pt}
   \noindent
   \if@twocolumn
     \bfseries\textit{\abstractname}: 
   \else
     \begin{center}\textbf{\abstractname}\end{center}
   \fi}
  {\par}
  
\renewenvironment{IEEEkeywords}
  {\normalfont
   \@IEEEabskeysecsize
   \vspace{0.3em}       
   \setlength{\parindent}{0pt}
   \noindent
   {\bfseries\itshape \IEEEkeywordsname: }\ignorespaces}
  {\par}

\def\IEEEbibitemsep{1.5pt plus .5pt}

\makeatother

\makeatletter
\AtBeginDocument{
  \renewcommand{\abstractname}{Abstract}
}
\renewcommand{\IEEEkeywordsname}{Keywords}
\makeatother

\begin{document}
\pagenumbering{gobble}
\fontsize{9}{10}
\selectfont

\fancyhead[RE,LO]{\centering{\small{\textit{Hội thảo khoa học Quốc gia về Trí tuệ nhân tạo 2026 (FJCAI) - Cần Thơ, 27-28/3/2026}}}}


%
\title{LatentFM: A Latent Flow Matching Approach for Generative Medical Image Segmentation}


\author{
    \IEEEauthorblockN{
        Ngoc Huynh Trinh,
        Hoang Anh Nguyen Kim,
        Hai Toan Nguyen,
        Quoc Long Tran
    }
    \IEEEauthorblockA{
        \textit{Institute for Artificial Intelligence, University of Engineering and Technology},\\
        \textit{Vietnam National University, Hanoi, Vietnam}\\
        Emails: \{huynhtn, nguyenhaitoan, tqlong\}@vnu.edu.vn,
    }
}

\maketitle
\begin{abstract}
Generative models have achieved remarkable progress with the emergence of flow matching (FM). It has demonstrated strong generative capabilities and attracted significant attention as a simulation-free flow-based framework capable of learning exact data densities. Motivated by these advances, we propose LatentFM, a flow-based model operating in the latent space for medical image segmentation. To model the data distribution, we first design two variational autoencoders (VAEs) to encode both medical images and their corresponding masks into a lower-dimensional latent space. We then estimate a conditional velocity field that guides the flow based on the input image. By sampling multiple latent representations, our method synthesizes diverse segmentation outputs whose pixel-wise variance reliably captures the underlying data distribution, enabling both highly accurate and uncertainty-aware predictions. Furthermore, we generate confidence maps that quantify the model’s certainty, providing clinicians with richer information for deeper analysis. We conduct experiments on two datasets, ISIC-2018 and CVC-ClinicDB, and compare our method with several prior baselines, including both deterministic and generative models. Through comprehensive evaluations, both qualitative and quantitative results show that our approach achieves superior segmentation accuracy while remaining highly efficient in the latent space.
\end{abstract}


\IEEEpeerreviewmaketitle

\section{Introduction}
Medical image segmentation is a crucial step in diagnosis, treatment planning, and image-guided surgery. It enables clinicians and medical experts to clearly identify and precisely localize abnormal regions such as lesions, tumors, and other pathological structures~\cite{litjens2017survey}. Manual segmentation, however, is labor-intensive, time-consuming, and increasingly impractical, motivating the need for automated solutions. With the rapid development of artificial intelligence, a wide range of Deep Learning (DL) models has been applied to medical image segmentation, ranging from convolutional neural network (CNN)-based architectures~\cite{ronneberger2015u, zhou2018unet++, isensee2021nnu, diakogiannis2020resunet} to more recent transformer-based models~\cite{chen2021transunet, cao2022swin, xie2021segformer, hatamizadeh2022unetr}. These models typically rely on labeled medical datasets and learn a direct mapping from the input image to its corresponding segmentation mask. Despite these advances, they still struggle with the inherent challenges of medical data, where anatomical structures are often ambiguous and boundaries can be difficult to delineate. As a result, deterministic approaches that produce only a single output become unreliable, fail to capture uncertainty, and ultimately limit overall performance~\cite{baumgartner2019phiseg}.

Recently, a new direction has emerged that focuses on learning the underlying data distribution rather than predicting a single deterministic mask. By modeling this distribution, generative approaches can capture richer structural information and produce diverse segmentation outputs. This ability yields more stable and uncertainty-aware predictions, which is crucial in medical imaging where ambiguity and inter-observer variability are common. Although originally developed for image synthesis, generative models have shown strong capacity to learn complex patterns, making them well-suited for segmentation tasks involving noisy or inherently ambiguous data. By sampling multiple segmentation candidates, these models not only express uncertainty but also provide outputs that are often more clinically meaningful. Various generative frameworks have been explored for this purpose, including VAE-based models~\cite{kingma2019introduction, kohl2018probabilistic, baumgartner2019phiseg}, generative adversarial network (GAN)-based models~\cite{goodfellow2014generative, xue2018segan} and diffusion-based models~\cite{ho2020denoising, amit2021segdiff, wu2024medsegdiff, wu2024medsegdiffv2}. More recently, the FM model~\cite{lipman2022flow} has also been adapted for this problem~\cite{bogensperger2025flowsdf, nguyen2025aleatoric}. FM has attracted significant attention due to its ability to learn exact data densities instead of relying on an evidence lower bound (ELBO), providing a promising direction for more accurate and robust segmentation.

In this paper, we further advance FM by introducing LatentFM, a novel flow-based model for medical image segmentation in the latent space. Our method integrates two VAEs to encode both images and masks into a shared latent space, combined with a conditional FM framework that enables the model to segment in this latent space. This design allows our model to effectively learn the underlying mask distribution while sampling multiple times to generate diverse segmentation masks for each input image. These samples can then be ensemble-averaged, producing more stable and reliable predictions compared with previous one-shot approaches. Furthermore, we also provide confidence maps that highlight regions where the model is more or less certain, offering clinicians valuable information for further analysis of detected anatomical or pathological regions.

\section{Related Work} \label{sec:related_work}
\textbf{Deterministic approaches:} Deep learning models for medical image segmentation have traditionally followed a deterministic paradigm, where each input image is mapped to a single segmentation mask. Representative CNN-based architectures include encoder–decoder networks such as UNet~\cite{ronneberger2015u}, which first introduced skip connections to preserve spatial information; UNet++~\cite{zhou2018unet++}, which enhances feature fusion via dense nested skip pathways; nnUNet~\cite{isensee2021nnu}, which automatically configures the network architecture and training pipeline; and ResUNet~\cite{diakogiannis2020resunet}, which incorporates residual connections to improve gradient flow and feature learning. In addition, transformer-based variants have recently shown strong performance, including TransUNet~\cite{chen2021transunet}, SwinUNet~\cite{cao2022swin}, SegFormer~\cite{xie2021segformer}, and UNETR~\cite{hatamizadeh2022unetr}. TransUNet~\cite{chen2021transunet} combines a vision transformer encoder with a UNet decoder to capture long-range dependencies while preserving local details. SwinUNet~\cite{cao2022swin} leverages hierarchical Swin Transformer blocks for efficient multi-scale feature extraction. SegFormer~\cite{xie2021segformer} employs a lightweight transformer backbone with MLP decoders, achieving strong segmentation performance with fewer parameters. UNETR~\cite{hatamizadeh2022unetr} uses a pure transformer encoder connected to a decoder via skip connections, providing effective volumetric segmentation, especially for 3D medical images. Overall, these models consistently demonstrate improved boundary reconstruction, multi-scale feature representation, and high accuracy across diverse medical imaging datasets. However, deterministic approaches limit the model to producing a single segmentation result for each input image. Medical imaging data often exhibit substantial variations in acquisition devices, noise levels, and pathological characteristics. Consequently, a single output may fail to capture the full range of plausible interpretations. In clinical practice, diagnoses are typically formed through consensus among multiple experts rather than relying on a single viewpoint. This motivates the development of generative approaches, which can produce multiple plausible segmentation outcomes and more effectively represent uncertainty in medical data.

\textbf{Generative approaches: } Unlike deterministic models, generative approaches can produce multiple outputs for each input image, thereby capturing a wider range of plausible anatomical variations. Early probabilistic methods include VAE-based architectures such as ProbUNet~\cite{kohl2018probabilistic}, which models segmentation uncertainty through a conditional latent space, and PHiSeg~\cite{baumgartner2019phiseg}, which introduces a hierarchical VAE to better capture multiscale variability. GAN-based models, such as SegGAN, have also been explored to improve boundary realism and handle complex structural details. In addition, diffusion-based models have demonstrated superior performance in recent studies, beginning with SegDiff~\cite{amit2021segdiff}, one of the earliest works applying diffusion models (DMs) to general image segmentation, followed by MedSegDiff~\cite{wu2024medsegdiff}, the first DM model specifically designed for medical image segmentation. Subsequently, the enhanced MedSegDiffv2~\cite{wu2024medsegdiffv2} has further improved stability and segmentation fidelity through refined diffusion processes and architectural optimizations. However, these previous generative models still face notable limitations in learning and estimating the underlying data distribution. GAN-based approaches often suffer from instability and mode collapse, making training unreliable, while VAEs and DMs rely on optimizing a variational bound on the likelihood, which provides only an indirect approximation of the true data distribution. Most recently, FM has emerged as a strong competitor to DM by directly approximating the likelihood function and enabling more stable and efficient generative modeling across various domains, including medical image segmentation. Specifically, FlowSDF~\cite{bogensperger2025flowsdf} leverages flow-based generative modeling to enhance segmentation quality, while another line of work focuses on estimating aleatoric uncertainty using FM formulations~\cite{nguyen2025aleatoric}, offering more reliable and interpretable predictions in clinical settings. In this paper, we further advance the FM framework by introducing a latent space formulation tailored for medical image segmentation, enabling more stable learning of the conditional mask distribution and more diverse segmentation outputs.

\section{Method} \label{sec:method}
\begin{figure*}[htbp]
    \centering
    \includegraphics[width=\textwidth]{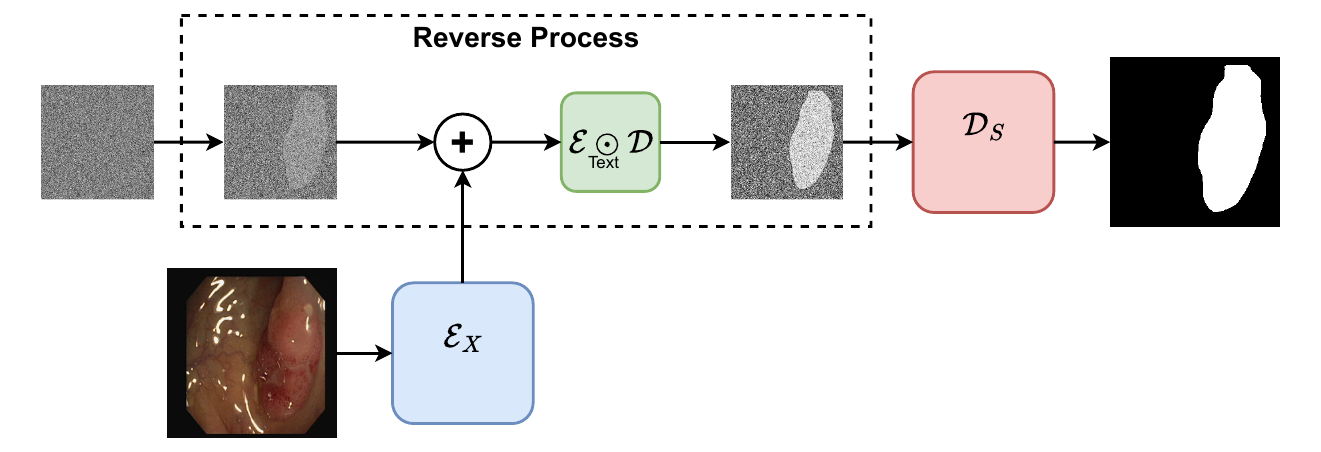} 
    \caption{Illustration of our pipeline for conditional FM-based segmentation in latent space during the reverse process.}
    \label{fig:pipeline}
\end{figure*}

In this section, we present LatentFM, a conditional FM framework operating in the latent space for medical image segmentation. We first introduce conditional FM and describe how it is adapted for the segmentation task. Next, we design two VAEs to encode images and masks into their respective latent spaces, while also enabling the decoding of the model’s outputs. Finally, we integrate these components to learn the conditional mask distribution and perform segmentation indirectly through the latent domain, as shown in Fig.~\ref{fig:pipeline}.

\subsection{Segmentation via Flow-matching}\label{cfm}
FM is a deep generative modeling framework originally developed for high-fidelity image synthesis. Unlike traditional generative models that rely on optimizing an ELBO loss function, FM directly maximizes the data likelihood by learning a continuous velocity field that transports a simple
prior distribution toward the data distribution. This formulation enables FM to better capture fine-grained structures and approximate the true data density more accurately. For the medical image segmentation task, we consider a labeled dataset:
\begin{equation}
    \mathcal{D} = \{ (X^{(i)}, S^{(i)}) \}_{i=1}^N ,
\end{equation}
where \(X^{(i)} \in \mathbb{R}^{H \times W \times C}\) denotes the \(i\)-th medical image, \(S^{(i)} \in \{0,1\}^{H \times W}\) is its corresponding segmentation mask, and \(N\) is the total number of image-mask pairs.

To model the conditional segmentation mask distribution \( q(S \mid X) \), we adapt the FM framework~\cite{lipman2022flow,lipman2024flow} to learn a conditional generative process that produces masks given the input image \(X\). Specifically, we construct a time-dependent probability path \(\{\, p_t(S \mid X) \,\}_{t \in [0,1]}\) that transports a simple source distribution \( p_0(S) \) (e.g., an isotropic Gaussian) toward the target distribution \( p_1(S) = q(S \mid X) \). We also define a flow \( \psi_t \) that evolves a source sample \( S_0 \sim p_0 \) according to the ordinary differential equation:
\begin{equation}
    \frac{d}{dt} \psi_t(S_0) = u_\theta\!\left(t, \psi_t(S_0), X\right),\quad S_t=\psi_t(S_0),
\end{equation}
where \( u_\theta(t, S_t, X) \) is a learnable, time-dependent velocity field conditioned on the image \(X\). The start state \( S_0 = \psi_0(S_0) \) and the terminal state \( S_1 = \psi_1(S_0) \) are respectively distributed according to the source distribution \( p_0(S) \) and the target conditional distribution \( q(S \mid X) \). To learn this velocity field, we introduce a conditioning latent variable \( z \) along the probability path \( \{ p_t(S \mid X, z) \}_{t \in [0,1]} \). The marginal segmentation distribution is obtained by integrating over \( z \):
\begin{equation}
    p_t(S \mid X) = \int p_t(S \mid X, z)\, p(z \mid X)\, dz,
\end{equation}
Following the FM framework~\cite{lipman2022flow,lipman2024flow}, one convenient choice is to condition the velocity field on both the source and target points. In this setting, we define the latent variable \( z \) as:
\begin{equation}\label{eq:z}
    z = (S_0, S_1)
    \sim p(z \mid X) = p_0(S) \, q(S \mid X),
\end{equation}
where \( S_0 \sim \mathcal{N}(0, I) \) is the source sample and \( S_1 \) is the ground-truth segmentation mask associated with the input image \( X \). Among the possible probability paths, we adopt the concentrated Gaussian path that linearly interpolates between \( S_0 \) and \( S_1 \). For a fixed standard deviation \( \sigma \), the conditional probability path is defined as:
\begin{equation}
    p_t\!\left(S \mid X, z=(S_0, S_1)\right)
    = \mathcal{N}\!\left((1-t) S_0 + t S_1,\; \sigma^2 I\right).
\end{equation}
As \( \sigma \to 0 \), the Gaussian collapses to this Dirac path and satisfies the exact boundary constraints:
\begin{align}
    p_{t=0}\!\left(S \mid X, z\right) &=  p_0(S),  \\
    p_{t=1}\!\left(S \mid X, z\right) &= \delta(S - S_1).
\end{align}
with \( \delta \) denoting the Dirac delta distribution. In this formulation, the segmentation mask at time-step \( t \) is given by the straight line interpolation:
\begin{equation}
    S_t = (1-t)\, S_0 + t\, S_1,
\end{equation}
and learning the velocity field reduces to a regression problem with the ground-truth velocity:
\begin{equation}
    u(t, S_t, X) = S_1 - S_0.
\end{equation}
Accordingly, the loss function is defined as:
\begin{equation}
    \mathcal{L}=\mathbb{E}_{t,X,S_0,S_1}\left[\left\|u_{\theta}(t,S_t,X)-(S_1-S_0)\right\|^2\right],
\end{equation}
where \(t\sim\mathcal{U}(0,1)\), \(X\) is the input image drawn from the dataset, and \(S_0, S_1\) are drawn following Equation~\ref{eq:z}.

Once the conditional FM model is trained, a segmentation map \( S_1 \) is obtained by integrating the velocity field:
\begin{equation}\label{eq:ode_1}
    S_1 = \psi_1(S_0, X), 
    \quad
    S_0 \sim p_0(S).
\end{equation}
Furthermore, multiple segmentation samples can be generated by drawing different source samples \( \{ S_0^{(i)} \}_{i=1}^n \) and propagating each through the learned flow:
\begin{equation}\label{eq:ode_n}
    S_1^{(i)} = \psi_1(S_0^{(i)}, X), 
    \quad
    i = 1, \ldots, n.
\end{equation}
The variability among the set \( \{ S_1^{(i)} \}_{i=1}^n \) can be quantified using the pixel-wise variance, which provides a confidence map reflecting the uncertainty in the segmentation process. In addition, by thresholding the averaged predictions at \( 0.5 \), we obtain the ensemble segmentation mask, yielding a more stable and reliable final prediction.

\subsection{Image and Mask in the latent space}

We utilize the VAE~\cite{kingma2019introduction} framework to construct two encoder–decoder models for medical images and their corresponding masks. First, we obtain a compact latent representation of the medical images through the image VAE. The image encoder \(\mathcal{E}_{X}\) maps an input image \(X \in \mathbb{R}^{H \times W \times C}\) to a latent vector \(z_{X} = \mathcal{E}_{X}(X)\), while the decoder \(\mathcal{D}_{X}\) reconstructs the image \(X' = \mathcal{D}_{X}(z_{X})\). The latent vector \(z_{X} \in \mathbb{R}^{h \times w \times c}\) lies in a lower-dimensional space, where the encoder downsamples the resolution by a factor \(f = H/h = W/w\). The downsampling factor \(f\) is a hyperparameter chosen as \(f = 2^{m}\) with \(m \in \mathbb{N}\).

For details, we assume \(X \sim p(X)\). A latent-variable generative model defines a joint distribution over \((X, z_X)\), with prior \(p(z_X)\) and likelihood \(p_{\theta}(X \mid z_X)\). The maximum marginal likelihood objective is:
\begin{equation}
    \mathbb{E}_{p(X)}[\log p_{\theta}(X)]
    = \mathbb{E}_{p(X)}\!\left[
        \log \mathbb{E}_{p(z_X)}
        \big[p_{\theta}(X \mid z_X)\big]
      \right].
\end{equation}
Since the true posterior \(p_{\theta}(z_X \mid X)\) is intractable, we introduce a variational approximation \(q_{\phi}(z_X \mid X)\), leading to the ELBO loss function:
\begin{equation}
\begin{aligned}
    \mathcal{L}_{\textrm{ELBO}}(X)
        &= \mathbb{E}_{q_{\phi}(z_X \mid X)}
           [\log p_{\theta}(X \mid z_X)] \\
        &\quad - D_{\textrm{KL}}\!\big(
                q_{\phi}(z_X \mid X)
                \,\|\, 
                p(z_X)
            \big).
\end{aligned}
\end{equation}
The first and second terms correspond to the reconstruction term \(\mathcal{L}_{\textrm{rec}}(\cdot)\) and the regularization term \(\mathcal{L}_{\textrm{reg}}(\cdot)\), respectively. Taking the expectation over the data distribution \(p(X)\) yields:
\begin{equation}
    \mathcal{L}_{\textrm{ELBO}}
    = \mathbb{E}_{p(X)}[\mathcal{L}_{\textrm{rec}}(X)]
    - \mathbb{E}_{p(X)}[\mathcal{L}_{\textrm{reg}}(X)].
\end{equation}
While this VAE models the image distribution \(p(X)\), we also require a compact latent representation for segmentation masks. Therefore, we use another VAE with encoder \(\mathcal{E}_{S}\) and decoder \(\mathcal{D}_{S}\), mapping a mask \(S\) to a latent code \(z_{S} = \mathcal{E}_{S}(S)\) and  reconstructing it via \(\mathcal{D}_{S}(z_{S})\). The mask VAE is also trained with the analogous ELBO, which provides a smooth, structured latent space for masks. These latent variables are subsequently used in our proposed LatentFM model, described in the next section.

\subsection{Segmentation via Latent Flow-matching}

With the two pretrained VAE models, the medical image and its corresponding segmentation mask are mapped into their respective latent spaces, yielding latent codes \(z_X=\mathcal{E}_X(X)\) and \(z_S=\mathcal{E}_S(S)\). In the latent domain, we incorporate the conditional FM framework to model the conditional latent mask distribution \(q(z_S \mid z_X)\). Formally, we learn a velocity field \(u_{\theta}(t, z_t, z_X)\), following the setup in Section~\ref{cfm}, which drives a latent flow \(z_t = \psi_t(z_0)\) from a source latent sample \(z_0 \sim p_0(z)\) toward the target latent mask \(z_1 = z_S\). Similar to the formulation in the image space, we choose a linear path to interpolate between the source sample \(z_0\) and the target sample \(z_S\). As \(\sigma \to 0\), the Gaussian path becomes increasingly concentrated and collapses to a Dirac distribution, thereby satisfying the exact boundary conditions. In our latent formulation, the trajectory at time \(t\) is therefore given by the straight-line interpolation:
\begin{equation}
    z_t = (1-t)\, z_0 + t\, z_S.
\end{equation}
This flow transports samples drawn from the simple prior distribution \(p_0(z)\) to the conditional latent mask distribution \(q(z_S \mid z_X)\), enabling the model to generate mask embeddings that are aligned with the latent image representation \(z_X\). Learning this velocity field also reduces to a regression problem with the ground-truth velocity:
\begin{equation}
    u(t, z_t, z_X) = z_S - z_0.
\end{equation}
Accordingly, our loss function is defined as:
\begin{equation}
    \mathcal{L} = \mathbb{E}_{t,\, z_X,\, z_0,\, z_S}\big[\|u_{\theta}(t, z_t, z_X) - (z_S - z_0)\|^{2}\big],
\end{equation}
where \(t \sim \mathcal{U}(0,1)\), \(z_X=\mathcal{E}_X(X)\) is the image latent code, \( z_0 \sim p_0(z) \) denotes a source latent sample drawn from the prior distribution, and 
\( z_S = \mathcal{E}_S(S) \) represents the latent code of the mask \( S \) corresponding to the ground-truth segmentation of image \( X \). Similarly, by reversing the process in Eqs.~\ref{eq:ode_1} and \ref{eq:ode_n}, we can sample output segmentation masks for each image. In addition, we incorporate the mask decoder \(\mathcal{D}_S\) at the end of the pipeline to decode the predicted latent representation back to the original mask space. The entire segmentation pipeline is illustrated in Fig.~\ref{fig:pipeline}, where we can iteratively sample multiple segmentation outputs for a given medical image. These samples are then used to produce a confidence map and a more stable and reliable averaged output mask.

\section{Experiments} \label{sec:experiments}

\subsection{Datasets} \label{subsec:dataset}

We conducted experiments on two medical image datasets. The first dataset was ISIC-2018, a large-scale collection of dermoscopic images from the International Skin Imaging Collaboration challenge, focusing on melanoma detection and segmentation. We used 3,694 paired dermoscopic images and masks, split into 3,194 for training, 250 for validation, and 250 for testing. The second dataset was CVC-ClinicDB, a widely used benchmark for polyp segmentation in colonoscopy images, containing 612 image-mask pairs. We divided this dataset into 492 training samples, 60 validation samples, and 60 testing samples. For both datasets, we resized all images to a spatial resolution of \(256 \times 256\) pixels and normalized them to the range \([-1, 1]\) using a standard preprocessing scheme with mean 0.5 and standard deviation 0.5, ensuring consistent input dimensions and stable optimization during training.

\subsection{Implementation Details} \label{subsec:implementation_detail}

We implemented two main types of models: two VAE models and a conditional FM model. First, we employed a VAE to encode image data from the original image space into a lower-dimensional latent representation. In both the ISIC-2018 and CVC-ClinicDB datasets, the input images were RGB images. Therefore, all inputs were represented as \(256\times 256\times3\), and the corresponding segmentation masks were represented as $256\times256\times1$. The VAE used a base channel size of 64 with channel multipliers \([1,2,4]\), and each stage consisted of residual and attention blocks. After encoding, both the image and its corresponding mask were projected into a latent space of size \(64\times 64\times3\). We trained the VAE for 250 epochs with a learning rate of \(1 \times 10^{-5}\) and a batch size of 4. Building on these latent embeddings, we implemented a conditional latent flow-matching model using a UNet architecture with a base channel size of 64 and channel multipliers \([1,2,2,4]\). Each stage again incorporated residual and attention blocks. We trained the model for 250 epochs with a fixed learning rate of \(1 \times 10^{-5}\). During inference, we performed five sampling iterations to simulate the decision variability of five clinicians, following the same protocol as MedSegDiff~\cite{wu2024medsegdiff}. We applied all hyperparameters consistently across both datasets.

\subsection{Results} \label{subsec:main_results}

\begin{figure*}[htbp]
    \centering
    \includegraphics[width=0.75\textwidth]{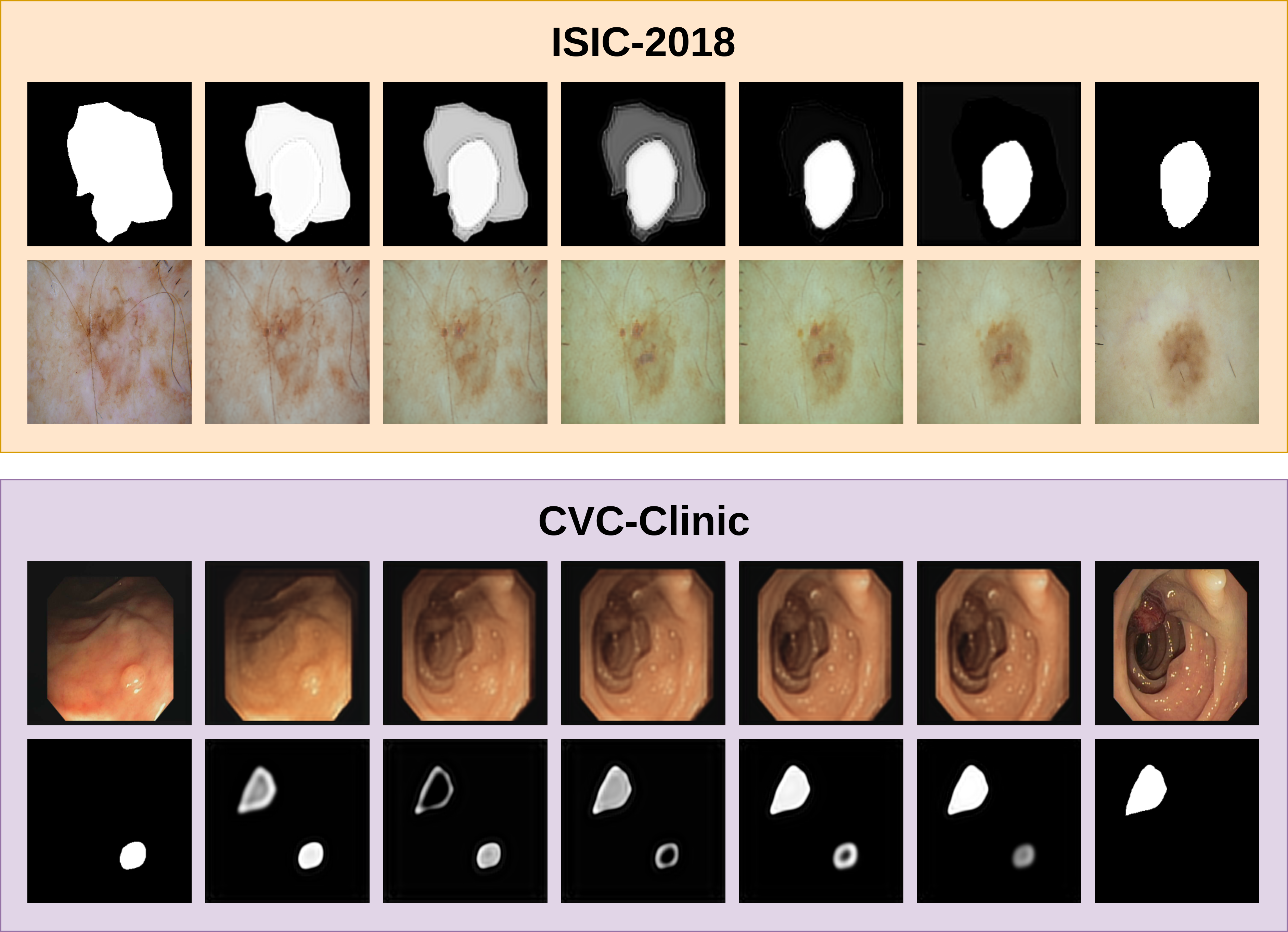}
    \caption{Illustration of interpolation between two images and their corresponding masks across two datasets.}
    \label{fig:interpolation}
\end{figure*}

We evaluated the two main components of our framework: the VAE models for both images and masks, and the conditional FM model for latent-space segmentation. The VAE results demonstrated stable and reliable reconstruction performance, as shown in Tab.~\ref{tab:reconstruction}. For image reconstruction, we used the Structural Similarity Index Measure (SSIM)~\cite{wang2004image} and Peak Signal-to-Noise Ratio (PSNR)~\cite{huynh2008scope} to quantify reconstruction quality. The VAE achieved an SSIM of 0.88 and a PSNR of 32.84 on ISIC-2018, and an SSIM of 0.90 with a PSNR of 33.56 on CVC-ClinicDB. These results indicated that the latent representations extracted from images were sufficiently informative to guide the downstream segmentation process. For mask reconstruction, we used two additional metrics, Dice and Intersection over Union (IoU), to evaluate performance. Our VAE consistently produced highly accurate results, reaching Dice and IoU scores of 0.99 on both datasets, along with SSIM scores above 0.98 and PSNR values above 34. These strong reconstruction metrics confirmed that the mask distributions were easier to model and that the learned latent space effectively captured the essential structural information. Overall, the VAE provided a stable latent embedding that facilitated effective encoding and decoding for the segmentation outputs generated by the latent FM model.

\begin{table}[htbp]
    \centering
    \caption{Reconstruction results of VAEs for both images and segmentation masks}
    \label{tab:reconstruction}
    \resizebox{\columnwidth}{!}{
    \begin{tabular}{lccccc}
        \toprule
        \textbf{Dataset} & \textbf{Input} & \textbf{Dice} & \textbf{IoU} & \textbf{SSIM} & \textbf{PSNR} \\
        \midrule
        \multirow{2}{*}{ISIC-2018}
        & Mask & 0.99 & 0.99 & 0.98 & 34.24 \\
        & Image & -- & -- & 0.88 & 32.84 \\
        \midrule
        \multirow{2}{*}{CVC-ClinicDB}
        & Mask & 0.99 & 0.99 & 0.99 & 36.04 \\
        & Image & -- & -- & 0.90 & 33.56 \\
        \bottomrule
    \end{tabular}
    }
\end{table}

Furthermore, we provided qualitative visual evidence demonstrating that the proposed VAE effectively models both images and masks, as shown in Fig.~\ref{fig:interpolation}. We performed latent space interpolation between two arbitrary images (or masks), whose latent representations were denoted as $z_1$ and $z_2$.  The intermediate latent codes were linearly interpolated and then decoded back into the image space. Specifically, the interpolation was formulated as:
\begin{equation}
    z_{\alpha} = \alpha z_1 + (1-\alpha) z_2,
\end{equation}
where $\alpha \in [0,1]$ controlled the interpolation ratio between the two latent representations. As observed in Fig.~\ref{fig:interpolation}, smooth transitions were achieved from the first sample to the second in both images and masks across the two datasets. This behavior suggested that the learned latent space was continuous and semantically meaningful, enabling the decoder to reconstruct plausible intermediate representations. Such continuity further indicated that the model could effectively decode novel latent vectors during the segmentation process. In addition, although we did not explicitly impose alignment constraints during training, a consistent structural correspondence between images and their masks was observed when performing interpolation in the shared latent space. These results demonstrated that the learned latent space preserved semantic consistency and structural alignment, thereby facilitating reliable conditional segmentation.

\begin{figure*}[htbp]
    \centering
    \includegraphics[width=\textwidth]{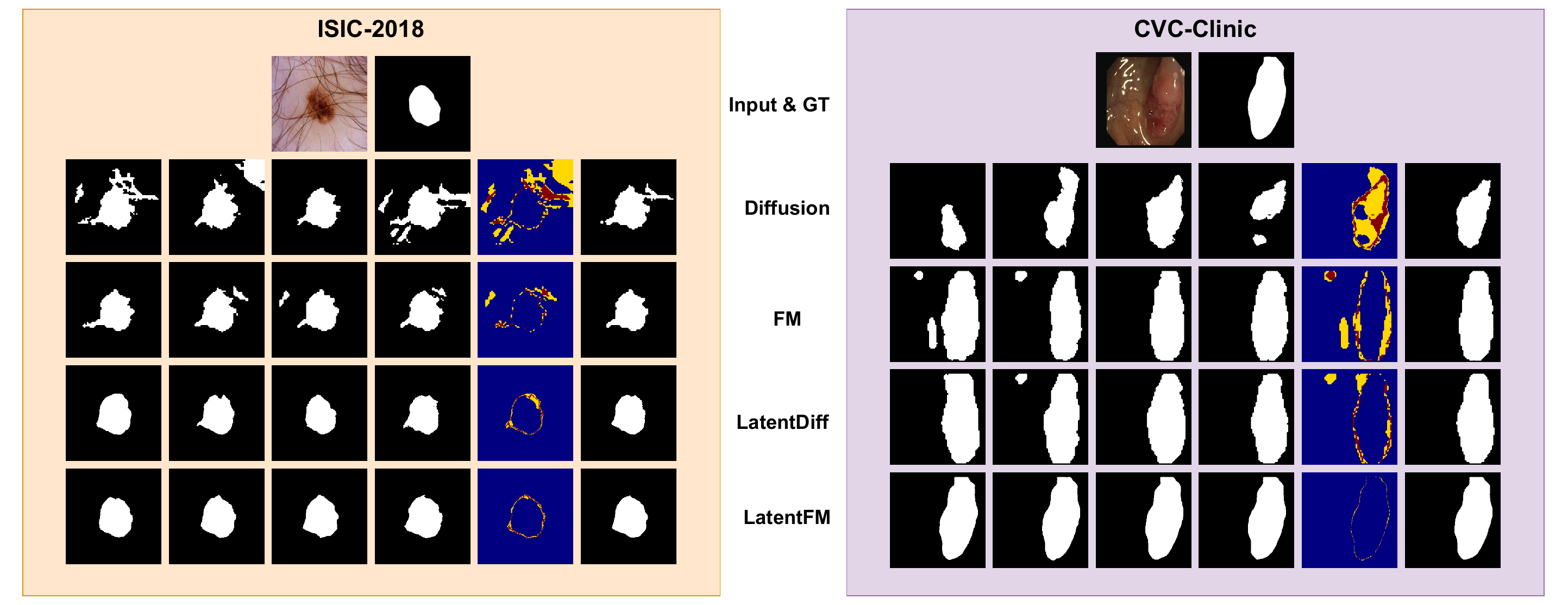}
    \caption{Qualitative comparison between our proposed LatentFM and other generative baseline approaches. For each model, the first four columns show the initial segmentation outputs, followed by the corresponding confidence map and the averaged output mask.}
    \label{fig:ensemble}
\end{figure*}

For the segmentation task, we compared our method with prior models from two major categories: deterministic and generative approaches, as shown in Tab.~\ref{tab:segmentation}. Overall, generative methods consistently outperformed deterministic baselines, yielding more accurate and reliable segmentation results. Among deterministic models, UNet and UNet++ achieved moderate performance and ranked lowest across datasets. More advanced architectures such as nnUNet and TransUNet achieved noticeably better performance. TransUNet exhibited stable results across both datasets, achieving Dice scores of 0.89 and 0.90 and IoU scores of 0.83 and 0.85 on ISIC-2018 and CVC-ClinicDB, respectively. nnUNet performed strongly on ISIC-2018 but showed weaker generalization on CVC-ClinicDB.

\begin{table}[b!]
    \centering
    \caption{Comparison of the proposed model with previous deterministic and generative approaches.}
    \label{tab:segmentation}
    \begin{tabular}{lcccc}
        \toprule
        \multirow{2}{*}{\textbf{Model}}
            & \multicolumn{2}{c}{\textbf{ISIC-2018}}
            & \multicolumn{2}{c}{\textbf{CVC-ClinicDB}} \\
        \cmidrule(lr){2-3} \cmidrule(lr){4-5}
            & Dice & IoU & Dice & IoU \\
        \midrule
        \multicolumn{5}{l}{\textbf{Deterministic Segmentation Methods}} \\
        \midrule
        UNet      & 0.8359 & 0.7550 & 0.8313 & 0.7469 \\
        UNet++    & 0.8100 & 0.7290 & 0.8016 & 0.7236 \\
        nnUNet    & 0.9080 & 0.8360 & 0.8130 & 0.7330 \\
        TransUNet & 0.8940 & 0.8220 & 0.9180 & 0.8590 \\
        \midrule
        \multicolumn{5}{l}{\textbf{Generative Segmentation Methods}} \\
        \midrule
        DM        & 0.8709 & 0.7714 & 0.8244 & 0.7313 \\
        LatentDM  & 0.9130 & 0.8410 & 0.8956 & 0.8111 \\
        \midrule
        FM        & 0.9101 & 0.8399 & 0.8902 & 0.8100 \\
        LatentFM  & \textbf{0.9511} & \textbf{0.9067}
                  & \textbf{0.9371} & \textbf{0.8816} \\
        \bottomrule
    \end{tabular}
\end{table}

In addition, we compared our method with generative models and provided both qualitative and quantitative results, as shown in Fig.~\ref{fig:ensemble} and Tab.~\ref{tab:segmentation}. The qualitative results indicated that DM struggled around mask boundaries and faced challenges in ambiguous cases, such as lesion regions with hair in the ISIC-2018 dataset. The subsequent models, FM and LatentDiff, produced more stable reconstructions with higher consistency across masks. The best performance was achieved by LatentFM, whose confidence map exhibited smooth variations that clearly reflected the model’s consensus and certainty. Aligned with the quantitative results, DM outperformed weaker deterministic baselines but remained slightly below TransUNet on average. LatentDM surpassed all deterministic models on ISIC-2018 with Dice 0.91 and IoU 0.84 but performed less strongly on CVC-ClinicDB. In contrast, FM-based models delivered notably stronger results than DMs due to their more accurate approximation of the target distribution. Our proposed LatentFM achieved the best overall performance, reaching Dice scores of 0.9511 and 0.9371 and IoU scores of 0.9067 and 0.8816 on ISIC-2018 and CVC-ClinicDB, respectively. Compared to LatentDM, LatentFM improved Dice by more than 0.04 and IoU by over 0.06 on both datasets. These results highlighted that the generative approach was better suited for medical image segmentation, and our proposed LatentFM, which modeled data distributions indirectly through the latent space, achieved the best performance and greater efficiency among methods following this direction.

\noindent\textbf{Limitations: } Despite the promising results, our study still has several limitations that warrant further investigation. First, the latent space resolution was adopted from prior frameworks, and a comprehensive analysis of different latent configurations has not been conducted.  A systematic comparison of resolution settings may provide further insights into the trade-off between representation capacity and computational efficiency. In addition, training and inference time remain important considerations for generative models, particularly for diffusion-based and flow-based approaches.  Although these models offer improved performance and uncertainty modeling, their computational cost may limit practical deployment in real-world clinical settings. Future work will focus on optimizing model efficiency and exploring lightweight variants to enhance scalability.

\section{Conclusion} \label{sec:conclusion}

In this paper, we addressed the medical image segmentation problem from a generative modeling perspective. Inspired by the strengths of FM, we extended and adapted this framework for segmentation tasks. Specifically, we designed a pair of VAEs to encode both medical images and masks into a compact latent space. Using these latent representations, our conditional FM module learned the latent mask distribution conditioned on the image latent code, effectively guiding the segmentation process. Consequently, the model generated multiple plausible segmentation masks for each input image, from which we derive confidence maps and ensemble-averaged predictions, resulting in more stable and reliable outputs. Experimental results showed that LatentFM consistently outperformed existing baselines, including both deterministic models and prior generative approaches. LatentFM achieved superior performance on both datasets while providing meaningful uncertainty estimation. In the future, we plan to further explore this direction by developing more comprehensive uncertainty quantification mechanisms, including explicit modeling of both epistemic and aleatoric uncertainty, to better address the inherent ambiguity in medical imaging tasks.

\bibliographystyle{IEEEtran}

\end{document}